%% file: AnoGen-CR.tex
\documentclass[runningheads]{llncs}

\usepackage[mobile]{eccv}

\usepackage{eccvabbrv}

\usepackage{graphicx}
\usepackage{booktabs}

\usepackage[accsupp]{axessibility}  % Improves PDF readability for those with disabilities.

\usepackage{algorithm}
\usepackage{algorithmic}
\usepackage{amssymb, amsfonts}
\usepackage{amsmath}
\usepackage{tensor}

\usepackage{multirow}
\usepackage{makecell}
\usepackage{color}
\usepackage{pifont}       % \ding{xx}
\usepackage{bbding}  
\usepackage[bold]{hhtensor}
\usepackage[misc]{ifsym}
\newcommand{\envelope}{\ding{41}}

\usepackage{wrapfig}

\newcommand{\squishlist}{
 \begin{list}{$\bullet$}
  { \setlength{\itemsep}{0pt}
     \setlength{\parsep}{1pt}
     \setlength{\topsep}{1pt}
     \setlength{\partopsep}{0pt}
     \setlength{\leftmargin}{1.5em}
     \setlength{\labelwidth}{1em}
     \setlength{\labelsep}{0.5em} } }
\newcommand{\squishend}{
  \end{list}  }
\usepackage{hyperref}

\usepackage{orcidlink}

\begin{document}

% ---------------------------------------------------------------
% TODO REVIEW: Replace with your title
\title{Few-Shot Anomaly-Driven Generation for Anomaly Classification and Segmentation} 

% TODO REVIEW: If the paper title is too long for the running head, you can set
% an abbreviated paper title here. If not, comment out.
\titlerunning{AnoGen}

% TODO FINAL: Replace with your author list. 
% Include the authors' OCRID for the camera-ready version, if at all possible.
% \author{First Author\inst{1}\orcidlink{0000-1111-2222-3333} \and
% Second Author\inst{2,3}\orcidlink{1111-2222-3333-4444} \and
% Third Author\inst{3}\orcidlink{2222--3333-4444-5555}}

\author{Guan Gui\orcidlink{0009-0004-5774-2555}~\inst{1,\star}\and
Bin-Bin Gao\orcidlink{0000-0003-2572-8156}~\inst{1,\star,\text{\envelope}} \\
Jun Liu~\inst{1} \and
Chengjie Wang\orcidlink{0000-0003-4216-8090}~\inst{1,2,\text{\envelope}} \and
Yunsheng Wu~\inst{1}
}

% TODO FINAL: Replace with an abbreviated list of authors.
\authorrunning{G. Gui and B.-B. Gao et al.}
% First names are abbreviated in the running head.
% If there are more than two authors, 'et al.' is used.

% TODO FINAL: Replace with your institution list.
% \institute{Princeton University, Princeton NJ 08544, USA \and
% Springer Heidelberg, Tiergartenstr.~17, 69121 Heidelberg, Germany
% \email{lncs@springer.com}\\
% \url{http://www.springer.com/gp/computer-science/lncs} \and
% ABC Institute, Rupert-Karls-University Heidelberg, Heidelberg, Germany\\
% \email{\{abc,lncs\}@uni-heidelberg.de}}

\institute{
Tencent YouTu Lab \and  Shanghai Jiao Tong University\\
\email{\{guiguan\}@smail.nju.edu.cn, 
\{csgaobb,junsenselee\}@gamil.com,
\{jasoncjwang,simonwu\}@tencent.com}
}

\maketitle
\let\thefootnote\relax\footnotetext{$^\star$ indicates equal contribution (G. Gui and B.-B. Gao). This research was done when G. Gui was an intern at YouTu Lab, Tencent, under the supervision of B.-B. Gao.}
\footnotetext{$^\text{\envelope}$ indicates corresponding authors.}

\input{sections/0_abstract}

\begin{figure}
    \centering
    \includegraphics[width=1.0\textwidth]{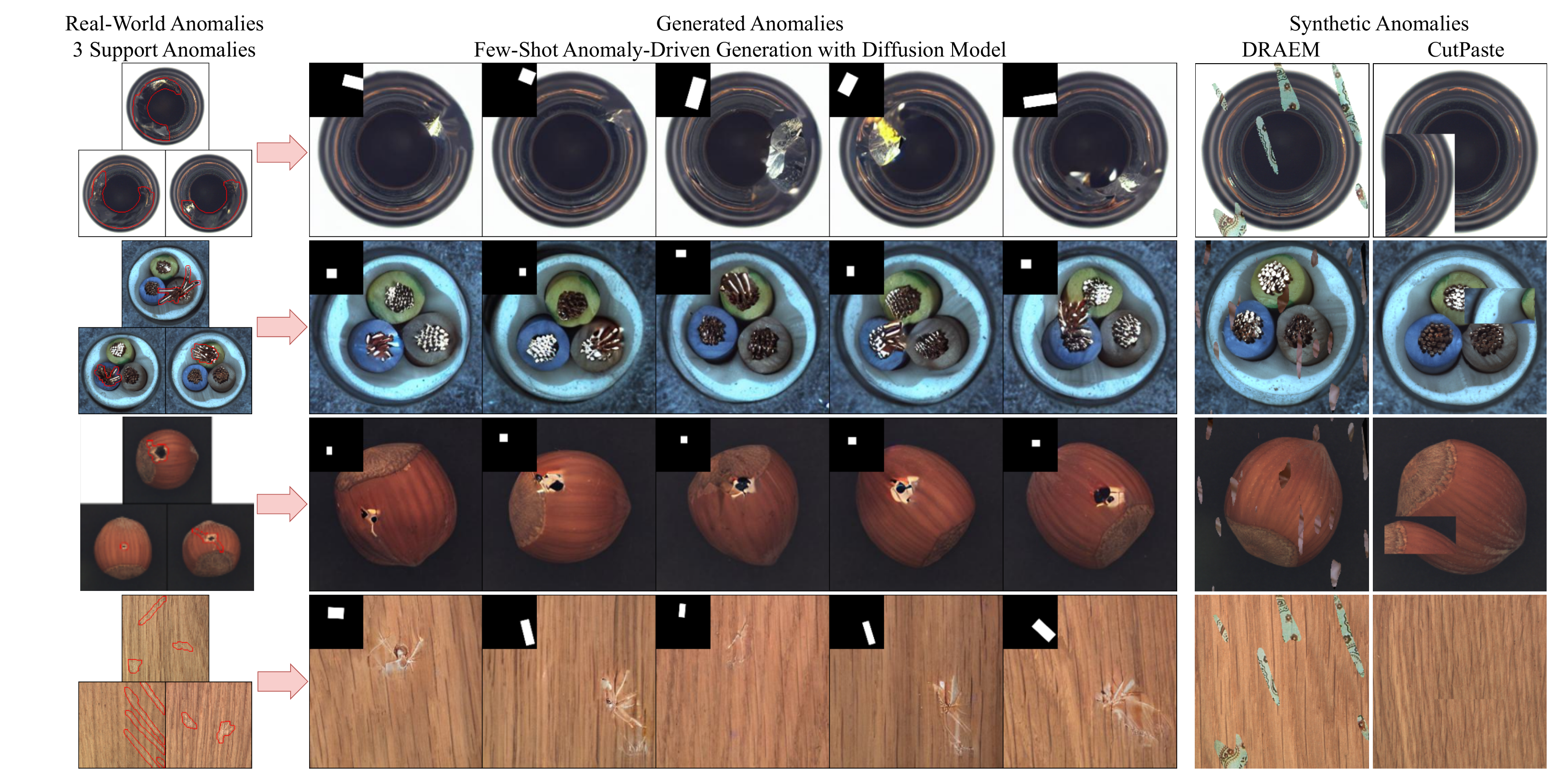}
    \caption{Comparisons of real anomalies (left column) and generated anomalies with ours (middle column) and other methods (right column). Given \emph{a few images of a real anomaly concept}, our AnoGen is able to generate more \emph{realistic and diverse anomalies} through learning a pre-trained diffusion model compared to the existing synthetic methods such as DRAEM and CutPaste. Meanwhile, our generated anomalies are spatially controllable because of a given mask (\eg, bounding box), which will benefit downstream anomaly detection tasks, \ie, classification and segmentation.}
    \label{fig:ga-vis}
\end{figure}

\input{sections/1_introduction}
\input{sections/2_related}

\input{sections/3_preliminary}
\input{sections/4_method}
\input{sections/5_experiments}
\input{sections/6_conclusion}

\bibliographystyle{splncs04}
\bibliography{anogen}

\end{document}

%% file: sections/0_abstract.tex
\begin{abstract}
     Anomaly detection is a practical and challenging task due to the scarcity of anomaly samples in industrial inspection. Some existing anomaly detection methods address this issue by synthesizing anomalies with noise or external data. However, there is always a large semantic gap between synthetic and real-world anomalies, resulting in weak performance in anomaly detection.
     To solve the problem, we propose a few-shot Anomaly-driven Generation (AnoGen) method, which guides the diffusion model to generate realistic and diverse anomalies with only a few real anomalies, thereby benefiting training anomaly detection models. Specifically, our work is divided into three stages. In the first stage, we learn the anomaly distribution based on a few given real anomalies and inject the learned knowledge into an embedding. In the second stage, we use the embedding and given bounding boxes to guide the diffusion model to generate realistic and diverse anomalies on specific objects (or textures). In the final stage, we propose a weakly-supervised anomaly detection method to train a more powerful model with generated anomalies. Our method builds upon DRAEM and DesTSeg as the foundation model and conducts experiments on the commonly used industrial anomaly detection dataset, MVTec. The experiments demonstrate that our generated anomalies effectively improve the model performance of both anomaly classification and segmentation tasks simultaneously, \eg, DRAEM and DseTSeg achieved a 5.8\% and 1.5\% improvement in AU-PR metric on segmentation task, respectively. The code and generated anomalous data are available at \url{https://github.com/gaobb/AnoGen}.
\end{abstract} 

%% file: sections/1_introduction.tex
\section{Introduction}
Anomaly detection has wide real-world application scenarios, \eg,  manufacturing quality inspection and medical out-of-distribution detection. However, the extreme scarcity of anomaly data in the real world makes anomaly detection tasks (including image-level classification and pixel-level segmentation) highly challenging.

Faced with the fact of rare anomaly data, several works propose unsupervised learning methods to eliminate the need for anomaly data.
For example, ~\cite{padim, dfm} estimate the multivariate Gaussian distribution of normal images, ~\cite{patchcore} creates a large memory bank to store the features of normal images, and ~\cite{gan1, riad, rd4ad} train a reconstruction network to compare the difference between reconstructed output and original image or feature extracted from a pre-trained model.
While these methods have achieved satisfactory performance in anomaly classification tasks, unfortunately, due to the lack of discriminative guidance from anomaly data, they still perform poorly in anomaly segmentation tasks.

To perform anomaly segmentation models better, some recent works, such as DRAEM~\cite{draem}, CutPaste~\cite{cutpaste}, and SimpleNet~\cite{simple}, propose to artificially synthesize anomalies to train discriminative models. Specifically, DRAEM mixes an external texture dataset and normal images to synthesize anomalies, CutPaste crops an image's region and randomly pastes it to another region, and Simplenet adds noise to the feature map to simulate anomalies. 
These synthesized anomalies have indeed proven beneficial for discriminative models, leading to superior performance in anomaly segmentation tasks. 
However, the drawback is that the synthesized anomalies are based on additional datasets or noise, which results in a significant semantic gap compared to real anomalies. This raises the question: \emph{is it possible to create realistic and diverse anomaly images that are semantically consistent with real-world anomaly concepts, thereby further enhancing these discriminative models?}

Fortunately, generative models have indeed achieved remarkable progress in creating realistic and diverse images. GANs~\cite{survy_gan} are trained to generate images by pitting a generative network against a discriminative network in an adversarial fashion. Although GANs are efficient in generating images with good perceptual quality, they are difficult to optimize and capture the full data distribution. Recently, Diffusion Models (DM)~\cite{survy_dm} have further surpassed GANs in image generation, which learn the distribution of images through a process of addnoising and denoising. However, these powerful generative model still requires a large number of training images, which raises the question: \emph{how to generate realistic and diverse anomaly images with the DM only a few real anomaly images are available?}

To solve the above problems, we propose a few-shot anomaly-driven generation method, which aims to generate realistic and diverse anomalies under the guidance of only a few real-world anomalies. Borrowing concepts in few-shot learning~\cite{fewshot}, we call these real-world anomalies ``support anomalies''. Considering that the number of support anomalies is very limited (1 or 3), it is not possible to optimize millions of parameters in the DM. Instead, we use a pre-trained diffusion model and then freeze its all parameters and only optimize an embedding vector that contains only a few hundred parameters. After training, this embedding vector is able to represent the distribution of given support anomalies, which guides the diffusion model to generate realistic and diverse anomalies as shown in Figure~\ref{fig:ga-vis}.

In the generation process, we provide a mask (a bounding box) condition to control the position and size of the anomaly region.
This mask also serves as a ground truth for downstream anomaly detection tasks, \ie, discriminative anomaly segmentation. Previous work~\cite{duan2023few} attempted to generate anomalies with GANs, while failing on downstream tasks due to the absence of labels. Specifically, we employ the generated images to two discriminative models: DRAEM~\cite{draem} and DeSTSeg~\cite{detseg}. Instead of using accurate masks in DRAEM and DeSTSeg, we have to take bounding box masks as supervision for training DRAEM and DeSTSeg. To this end, we propose a weakly-supervised learning~\cite{weakly} version built on DRAEM and DeSTSeg, where high-confidence normal predictions within the box region will be filtered out to alleviate their interference for model training.

It is worth noting that a concurrent study, AnomalyDiffusion~\cite{hu2024anomalydiffusion}, shares similar concepts to ours. It also generates more anomalies with a small number of real anomalies, but their implementations are different. First, AnomalyDiffusion is more complex, it trains a mask generation network, which significantly increases computational costs. In contrast, we only learn a 768-parameter embedding. Then, AnomalyDiffusion utilizes prior knowledge of anomaly masks to constrain its generated shape. This potentially limits the diversity of generated anomalies. While we do not impose such constraints and thus retain the diversity of generated anomaly. Finally, to alleviate the problem of inaccurate masks, AnomalyDiffusion uses adaptive attention re-weighting to fill the mask region. However, it still cannot completely solve it. We propose a weak supervision method, which is more effective in addressing this issue, making our approach more robust and generalizable.

We conduct experiments on the commonly used industrial anomaly detection dataset, MVTec~\cite{bergmann2019mvtec}. With the help of our generated images equipped with the proposed weakly supervised anomaly detection method, we successfully improved the anomaly detection performance of both DRAEM (from 67.4\% to 76.6\% in P-AUPR)  and DeSTSeg (from 73.2\% to 78.1\% in P-AUPR) models. In a word, our contributions can be summarized as:

\squishlist 
    \item We propose a few-shot anomaly-driven generation method guiding a diffusion model to generate realistic and diverse anomalies with a few real-world anomalies. These generated anomalies are consistent with real-world anomalies in semantics.
    \item We propose a bounding-box-guided anomaly generation process, which not only allows for control over the position and size of the anomaly regions but also provides bounding box supervision for discriminative anomaly detection models.
    \item Based on bounding box supervision, we propose a simple weakly-supervised anomaly detection method built on two discriminative anomaly models, DRAEM and DeSTSeg. The experiments on MVTec show that we successfully improve their performance both on anomaly classification and segmentation tasks.
\squishend 

%% file: sections/2_related.tex
\section{Related Work}
\noindent\textbf{Conditional Diffusion Models.}
The diffusion model~\cite{diffusion} has achieved remarkable success in image generation, leading to a surge of research interest in generating images that meet specific user expectations under given conditions.
For instance, ~\cite{class-diffusion} focuses on generating images corresponding to class labels.
Other methods such as ~\cite{nichol2021glide, more-control, imgaen} allow users to provide text and generate images that align with the provided text. ~\cite{singh2022conditioning, choi2021ilvr} utilize reference images to generate images with similar style and structure. 
~\cite{li2023gligen, chai2023layoutdm} refer to the given layout conditions, which ensure that the elements in the image have the expected relative positional structure.
Additionally, ~\cite{meng2021sdedit, balaji2022ediffi} enables users to generate corresponding images based on the user's sketches.
However, these methods are mainly applied to natural images and generate semantically correct images without being able to generate specific objects.

There is a huge gap between industrial anomalies and natural images. In this paper, we aim to generate special anomalies based on a given anomaly concept. In~\cite{shi2023instantbooth, ruiz2023dreambooth}, they can generate a specific concept with given reference images, but they focus on generating the entire object. However, the industrial anomalies are often only a small part of the object. In addition, to control the specific area of the generated anomalies, we use the inpainting mode, which has been explored in~\cite{inpaint1, blended}.

Furthermore, the use of diffusion models to generate images has proven to be beneficial for various downstream image recognition tasks. For example, ~\cite{diffusionaug} employs generated images as data augmentation, ~\cite{diffusionimagenet} generates images to enhance training ImageNet classification model, and~\cite{akrout2023diffusionskin} generates skin images for disease classification task.
\emph{As far as we know, we are the first to utilize diffusion models to generate images and assist in anomaly detection.}

\noindent\textbf{Industrial Anomaly Detection.}
Unsupervised anomaly detection models are generally based on embedding or reconstruction to learn normal distribution.
In embedding-based methods, ~\cite{padim, dfm} use a pre-trained network to extract features from images and estimate their multivariate Gaussian distribution. 
~\cite{cs-flow} fits the normalizing flow into a distribution in the form of a product of Gaussian and Dirac distributions.
~\cite{patchcore} maintains a huge memory bank to save the extracted features.
~\cite{reiss2021panda, CFA} propose feature adaptation for adapting targeted datasets.
Reconstruction-based methods~\cite{riad,sspcb} encourage a model to learn the masked areas.~\cite{mou2022rgi} proposes a robust GAN-inversion to store any images. ~\cite{uniad} uses a layer-wised query transformer to reconstruct original features preventing a shortcut issue.

Discriminator-based models often achieve superior performance in anomaly segmentation tasks and they require synthetic anomaly images to train the discriminator.
~\cite{draem} utilizes an external texture dataset to construct anomalies, while ~\cite{cutpaste} randomly crops image patches and pastes them onto other parts to create anomalous images. 
~\cite{simple} directly perturbs the feature map to simulate the features of anomalous images. 
However, these synthetic anomalous images have a significant distribution gap with real-world anomalous ones, which is not conducive to training anomaly detection models in practical applications. \emph{We expect the generated images to be as similar as possible to real anomalies, making the model more robust.}

%% file: sections/3_preliminary.tex
\section{Preliminaries}

\noindent\textbf{Image Generation with Diffusion Models.} The diffusion model (DM) is a probabilistic model for learning data distribution, which can reconstruct diverse samples from noise. The DM regards the process of addnosing and denosing as a Markov chain of length $T$, and learns the data distribution through a continuous process of addnosing and denoising. The optimization objective of DM is to predict noise from the noisy image, which can be expressed as:
\begin{equation}
    L_{DM} = \mathbb{E}_{x, \epsilon \sim \mathcal{N}(0,1),t}[||\epsilon-\epsilon_{\theta}(x_t,t)||^2_2],
\end{equation}
where $t$ uniformly sampled from $[1,\cdots, T]$, $\epsilon$ is the noise sampled by a Gaussian distribution, $x_t$ is the noisy version during the addnoising process, and $\epsilon_{\theta}(x_t,t)$ is the noise predicted by the network during the denoising process.

\noindent\textbf{Conditional Diffusion Models.} Given a random noise, DM can generate diverse images through iterative denoising. However, the semantics of the generated image are uncontrollable. 
To solve the problem, the Latent Diffusion Model (LDM) proposes to use a condition $y$ to control the denoising process. LDM first transforms the images into a latent space, then injects the condition $y$ into the model through a cross-attention module, allowing the model to generate images corresponding to $y$. The optimization objective of LDM can be simplified as:
\begin{equation}
    L_{LDM} = \mathbb{E}_{\varepsilon(x), \epsilon,t,y}[||\epsilon-\epsilon_{\theta}(\varepsilon(x),t,\tau_{\theta}(y))||^2_2],
    \label{eq2}
\end{equation}
$\varepsilon(\cdot)$ is a auto-encoder and is used to transform $x$ into the latent space. The condition $y$ could be a text, a class label, etc., and $\tau_{\theta}(\cdot)$ is the corresponding encoder.

Intuitively, one might think that training an expert model $\tau_{\theta}(\cdot)$ to encode the industrial prompt, such as ``scratch'', would be sufficient for generating anomaly images. However, this approach becomes impractical due to the scarcity of anomaly images. Similarly, fine-tuning the LDM is not feasible for the same reason.
To overcome these challenges and reduce the reliance on anomaly images, we propose a scheme that directly utilizes embeddings as conditions. By doing so, we only need to learn an embedding with a small number of parameters, typically a few hundred. This approach allows us to generate anomaly images effectively while mitigating the limitations imposed by the scarcity of anomaly data.

\noindent\textbf{Discriminative Anomaly Detection Model.}
A discriminative model typically consists of two modules: reconstruction and discrimination. The reconstruction module is responsible for reconstructing anomalous (normal) images into normal (itself) ones, and then the discrimination module predicts the segmentation map of the anomalous regions based on the difference between the reconstructed output and the original input.
To formalize the process, let's use DRAEM as an example to explain.
Assuming $I_n$ denotes a normal image, and the corresponding synthetic anomalous image is $I_a$. We denote the reconstructed output of $I_a$ or $I_n$ as $I_r$, then the optimization objective for reconstructive sub-network is
\begin{equation}
    L_{rec}=\lambda L_{SSIM}(I,I_r)+l_2(I,I_r),
\end{equation}
where $\lambda$ is a weight balancing between SSIM~\cite{SSIM} loss and $l_2$ loss.
Then, a normal image $I_n$ (or its synthetic version $I_a$) and the corresponding reconstruction version $I_r$ are considered as the input of the discriminative sub-network, and the optimization objective of the discriminative sub-network is
\begin{equation}
    L_{seg} = L_{Focal}(M, \hat{M}),
    \label{l_seg}
\end{equation}
where $\hat{M}$ is the predicted segmentation map, $M$ is the ground truth mask of $I_n$ (or $I_r$) and $L_{Focal}$ is the Focal Loss~\cite{focal}.

It can be seen that both reconstruction and discrimination modules require anomalous images $I_a$. In DRAEM and DeSTSeg, they create anomalies by blending normal images with an external dataset, DTD~\cite{texture}, while CutPaste creates anomalies by randomly pasting image patches. As mentioned earlier, these synthetic anomaly images are semantically inconsistent with real-world images. Therefore, our goal is to generate semantically consistent images to further enhance discriminator-based models.

%% file: sections/4_method.tex
\begin{figure*}[tb]
\centering
\includegraphics[width=1.0\textwidth]{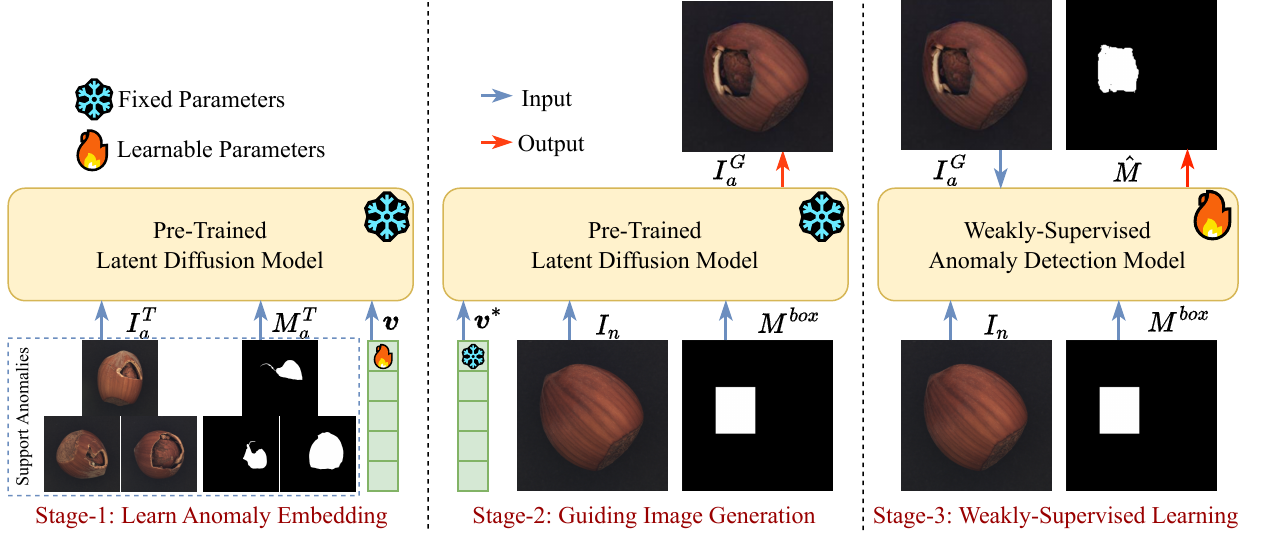} 
\caption{Pipeline of our work, and it consists of three stages. In the first stage, we learn an embedding vector $\boldsymbol{v}$ with few support anomalies ($I_a^T$, $M_a^T$) based on a pre-trained Latent Diffusion Model (LDM) fixing all parameters, where the number of real-world anomalous images $I_a^T$ is only 1 or 3, and $M_a^T$ is the corresponding ground-truth masks.
In the second stage, given a normal image $I_n$ and a bounding box mask $M^{box}$, we use the learned embedding $\boldsymbol{v}^*$ to guide the LDM to generate anomalous image $I_a^G$.
In the third stage, we use the normal image $I_n$, bounding box mask $M^{box}$, and generated image $I_a^G$ to train a weakly-supervised anomaly detection model for image-level classification and pixel-level segmentation.
}
\label{fig2}
\end{figure*}

\section{Methodology}
Our method consists of three stages, as shown in Figure~\ref{fig2}. The first stage learns embeddings based on a few support anomalies and their ground-truth segmentation maps. In the second stage, anomalies are generated by leveraging the learned embeddings, given objects (or textures) and bounding boxes as guidance. In the third stage, a weakly supervised anomaly detection model is trained using anomalies and bounding box supervision.

\subsection{Stage-1: Learn Anomaly Embedding}
When dealing with a limited number of anomalous images, it becomes impractical to optimize a diffusion model with millions of parameters. However, the optimization process becomes much easier when working with an embedding that consists of only a few hundred parameters. 
Hence, our choice is to focus on learning an embedding that effectively captures the semantic characteristics of real anomalies, rather than relying on fine-tuning a complex model.

Given that predicting noise in LDM involves learning the data distribution, we can leverage the loss associated with noise prediction to gain insights into the distribution of real anomalies.
Specifically, we first initialize an embedding $\vec{v}$ to replace the condition embedding $\tau_{\theta}(y)$ in LDM, and then optimize it with Equ~\ref{eq2}. It can be denoted as follows:
\begin{equation}
    \vec{v}^* = \mathop{\arg\min}\limits_{\vec{v}} L_{LDM}(I_a^T,t,\vec{v}),
\end{equation}
where $I_a^T$ is a few (\eg, 1 or 3) real-world anomalous images. Similar to the conditioning mechanism in LDM, we insert embedding $\vec v$ into intermediate layers of the UNet in LDM implementing with cross-attention. Instead of learning the entire network, we initiate the embedding and freeze the parameters of a pre-trained LDM model. 
This allows us to solely update the embedding during the addnoising and denoising process. 
By doing so, the learned embedding captures distribution about the provided real-world anomalies, which subsequently guides the image generation in the subsequent stage.

In certain cases, the anomaly region within an image is typically a small fraction of the overall object (\eg, ``\textit{a hole in a carpet}'').
Training the model on the entire image may result in a learned data distribution that is biased towards the object itself (\eg, ``\textit{carpet}'') rather than focusing on the anomaly (\eg, ``\textit{hole}'').
To address this concern, we incorporate the segmentation mask of the abnormal image to guide the loss function.
We denote $M_a^T$ as the segmentation mask of $I_a^T$, then the modified LDM loss can be described as
\begin{equation}
    L_{LDM}^{'}=\mathbb{E}_{\varepsilon(x), \epsilon,t,\vec{v}}[||(\epsilon-\epsilon_{\theta}(\varepsilon(I_a^T),t,\vec{v})) \odot M_a^T||^2_2],
\end{equation}

There are some similar works~\cite{imagic, textual} to this approach, where they also learn the specified object by optimizing an embedding. However, these works focus on learning the entire object, whereas our approach emphasizes capturing the local details of the object through a mask-guided loss. We demonstrate the impact of these two approaches on anomaly generation in the following experiments.

\subsection{Stage-2: Guiding Anomaly Generation}
Our generation objective is to ensure that the generated images exhibit both semantic and spatial controllability. On the semantic level, we aim to generate images that are consistent with real-world examples, maintaining consistency in terms of objects or textures (\eg \textit{``bottle"}) and the type of anomaly (\eg \textit{``broken"}).
On the spatial level, we strive to control the position and size of the anomaly region by providing bounding boxes. 
By achieving both semantic and spatial controllability, we can generate images that closely align with our desired specifications.

To achieve the above goals, we adopt an inpainting technique inspired by ~\cite{blended}.
Specifically, we randomly sample a normal image $I_n$ from the training set as the input image and employ a bounding box mask $M^{box}$ to regulate the location and size of the generated anomaly.
The embedding $\vec v^*$ will be frozen and injected into the image as a condition through the cross-attention module, thus generating the expected anomaly.
For the inference image at each step in the denoising process, the area within the box will be retained, and the area outside the box will be replaced by the noisy version of $I_n$.
By doing so, we can control the generated anomalies to be located in specified areas of the input image while leaving other areas untouched.
We can represent this process as:
\begin{equation}
    z_t = z_{n}^t \odot (1-M^{box}) + z_{t}^{'} \odot M^{box},
\end{equation}
where $z_{n}^t$ is the addnosing version of $\epsilon(I_n)$ at $t$ step, and $z_{t}^{'}$ is the denoising version of $z_{t+1}$. 
$\epsilon(\cdot)$ transforms $I_n$ into the latent space.
After the denoising process, the latent variable $z_0$ is passed through a decoder to produce anomalous image $I_a^G$.
Since the diffusion model generates images from random noise, they inherently possess a certain degree of diversity. However, to augment this diversity further, we introduce ground boxes with arbitrary positions and sizes.

\subsection{Stage-3: Weakly-Supervised Anomaly Detection}

Existing discriminative models are typically trained on precise masks. However, it is not suitable for our bounding box supervision since not all pixels within the box are necessarily anomalous. If we directly use the entire box as supervision for the anomaly region, it would mistakenly classify normal pixels as anomalous ones and thus damage the model performance in the anomaly segmentation task. 
To accommodate the bounding box supervision, we design a weakly supervised loss for anomaly detection.

Let $\hat p_{(i,j)}$ represent the predicted normal probability at the location $(i,j)$. We use a threshold $\tau$ to distinguish confident predictions:
\begin{equation}
	\delta_{(i,j)} = \begin{cases}
	      1, & \text{if} ~ \hat p_{(i,j)} \ge \tau \\
	      0, & \text{otherwise}
		   \end{cases},
\end{equation}
where $\hat p_{(i,j)}=1-\hat M_{(i,j)}$. For high-confidence normal pixels within the box region, we set their loss to 0, thereby reducing the confliction of possible normal pixels within the bounding box. For all pixels out of the bounding box, we use the same segmentation loss $L_{seg}$ as Eq.~\ref{l_seg}. Therefore, the overall weakly-supervised loss of discriminative sub-network is
\begin{equation}
    L_{seg}^{'}= M^{box} \odot (1-\delta) \odot L_{seg} + (1-M^{box}) \odot L_{seg},
\end{equation}
where $M^{box}$ is the given box mask in anomaly generation.

%% file: sections/5_experiments.tex
\section{Experiment}

\subsection{Implementation Details}
\noindent\textbf{Datasets}. MVTec~\cite{bergmann2019mvtec} is a widely used industrial anomaly detection dataset that contains 10 objects and 5 textures, each with 1-8 types of anomaly and a few anomalous images, totaling 73 types of anomalies and 1258 anomalous images.
For each type of anomaly, we generate 4 anomalous images for each object (or texture) in the training set, obtaining an anomaly dataset of 70,760 images.

\noindent\textbf{Learning embeddings and generated images}. We use the pre-trained LDM ~\cite{ldm} without any parameter fine-tuning. The model uses the text encoder in the CLIP~\cite{clip} to obtain text embeddings, thus we use the word ``defect'' through the text encoder to obtain the initialized embedding $\vec{v}$ (dimension is 768).
In the stage of learning $\vec{v}$, we randomly select 3 anomalous images from the real anomalies to be the support anomalies. We train 6000 iterations with a learning rate of 0.005.
During the stage of generating anomalous images, we randomly create 2 masks for each object (or texture) and generate 2 anomalous images using each mask.

\noindent\textbf{Bounding-Box Generation}.
In order to enhance the rationality of the generated anomalies, we impose certain constraints on the bounding box. Specifically, we control position to ensure at least 50\% IoU between the bounding box and the foreground region obtained with GrabCut\cite{rother2004grabcut}. Then, we set hyper-parameters to control the size, \eg, the hyper-parameters of hazelnut-hole are [0.1, 0.5], indicating that the bounding box is between 10\% and 50\% of the image width or height. Based on these above constraints, we randomly generate bounding boxes to enhance the diversity of generated anomalies.

\begin{figure*}[tb]
\centering
\includegraphics[width=1.0\textwidth]{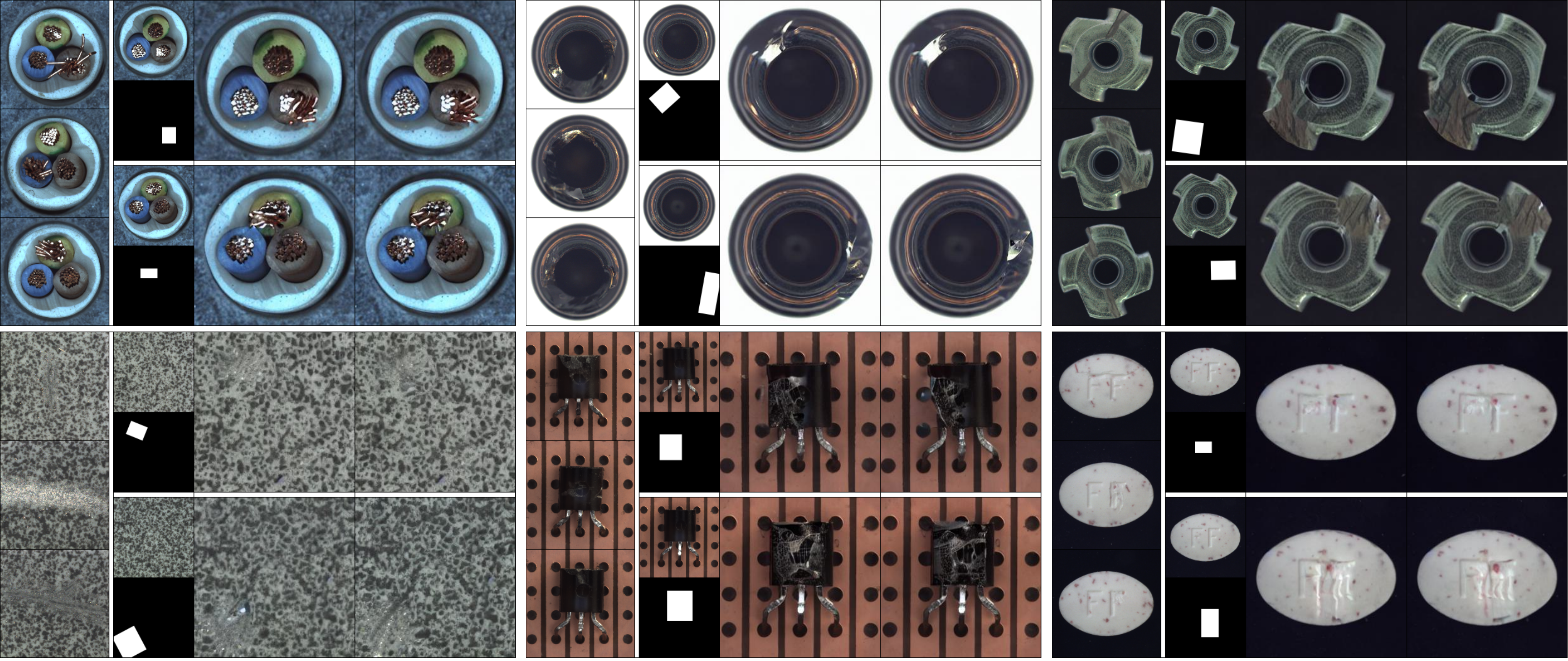} 
\caption{We show six sets of images, in each set, the first column is the support anomalies (only 3 images), and the second column is the object (or texture) sampled from the training set with a randomly generated bounding box mask, the third and fourth columns are the generated anomalous images.}
\label{fig3}
\end{figure*}

\begin{table*}[]\scriptsize
\centering
\tabcolsep=3pt
\caption{Anomaly classification comparisons (image-level AU-ROC / AU-PR) and anomaly segmentation comparisons (pixel-level AU-ROC / AU-PR) on MVTec. The results of DRAEM and DeSTSeg are reported by running their official code. To ensure a fair comparison, we ran DRAEM, DeSTSeg and our AnoGen under the same environment, while keeping the hyper-parameters consistent with the original paper.}
\begin{tabular}{cc|cccc|cccc}
\toprule
\multicolumn{2}{c|}{ Metrics}                 &       \tiny CS-FLow  & \tiny PaDim   & \tiny PatchCore   & \tiny RD4AD  & \tiny DRAEM & \multicolumn{1}{c|}{ \tiny \textbf{AnoGen}}                        & \tiny DeSTSeg& \tiny \textbf{AnoGen} \\ \midrule
\multicolumn{1}{c|}{}                        & \tiny AU-ROC & 97.5      & 91.2    & 97.8        & 98.7   & 97.1  & \multicolumn{1}{c|}{\textbf{98.7} {\textcolor{red}{(1.6~$\uparrow$)}}}                        & 98.3    & \textbf{98.8} {\textcolor{red}{(0.5~$\uparrow$)}} \\
\multicolumn{1}{c|}{\multirow{-2}{*}{\tiny Image}} & \tiny AU-PR  & 97.7      & 94.2    & 98.8        & 97.8   & 98.5  & \multicolumn{1}{c|}{\textbf{99.5} {\textcolor{red}{(1.0~$\uparrow$)}}}                        & 99.4    & \textbf{99.6} {\textcolor{red}{(0.2~$\uparrow$)}} \\ \midrule
\multicolumn{1}{c|}{}                        & \tiny AU-ROC & 93.4      & 96.9    & 97.5        & 93.9   & 96.8  & \multicolumn{1}{c|}{\textbf{98.1} {\textcolor{red}{(1.3~$\uparrow$)}}}                        & 98.2    & \textbf{98.8} {\textcolor{red}{(0.6~$\uparrow$)}} \\
\multicolumn{1}{c|}{\multirow{-2}{*}{\tiny Pixel}} & \tiny AU-PR  & 59.6      & 48.5    & 61.7        & 55.4   & 67.4  & \multicolumn{1}{c|}{\textbf{73.2} {\textcolor{red}{(5.8~$\uparrow$)}}}                        & 76.6    & \textbf{78.1} {\textcolor{red}{(1.5~$\uparrow$)}} \\ \bottomrule
\end{tabular}
\vspace{-5pt}
\label{tab1}
\end{table*}

\noindent\textbf{Anomaly detection task}. We use DRAEM and DeSTSeg as baseline models.
Apart from the hyper-parameter $\tau=0.9$, we keep the other hyper-parameters consistent with the original paper to ensure fair comparisons.
When sampling a training batch, we randomly sample the original synthetic anomalies and our generated anomalies with a probability of 0.5.
For evaluating the performance, we compared the AU-PR and AU-ROC metrics for both anomaly classification and anomaly segmentation tasks. It is worth noting that due to the severe imbalance between normal and anomalous pixels, the AU-PR metric on the segmentation task may better measure the performance of the model~\cite{roc}.
We focus on comparing with the DRAEM and DeSTSeg to validate that our generated anomalous images can benefit the model train better. In addition, we also compared with other unsupervised anomaly detection models: PaDim~\cite{padim}, PatchCore~\cite{patchcore}, CS-Flow~\cite{cs-flow} and RD4AD~\cite{rd4ad}.

\subsection{Qualitative Analysis for Anomaly Generation}
In Figure~\ref{fig3}, we present a selection of generated anomalies (Additional generated images are shown in the supplementary materials). 
Our generated anomalies effectively meet our expectations, despite the presence of only 3 real anomalies in the support set. They exhibit similarity to real-world anomalies and demonstrate diversity. 
Moreover, by applying box conditions, we gain control over the position and size of the anomaly region, enabling spatial controllability.

\subsection{Quantitive Comparisons on Anomaly Detection}
We evaluate the effectiveness of our generated images by assessing the model's performance on anomaly detection tasks. We focus on whether our generated images can directly improve the model's performance.
As shown in Table~\ref{tab1}, both DRAEM and DeSTSeg consistently exhibit improvements across all metrics.
For example, we observe a 1.6\% enhancement in image-level AU-ROC for DRAEM and a 0.5\% improvement for DeSTSeg. Similarly, in terms of pixel-level AU-PR, DRAEM and DeSTSeg show improvements of 5.8\% and 1.5\% respectively.
This demonstrates that incorporating real-world anomaly distribution guidance can be beneficial for the model, especially in the anomaly segmentation task.
Then, when compared to unsupervised learning models, it becomes evident that discriminator-based models outperform in anomaly segmentation tasks. This finding reveals the guidance of anomalies is of great significance for anomaly segmentation tasks.
These conclusions strongly support the crucial importance of providing anomalies that are consistent with real-world anomalies for the performance of anomaly segmentation models, which aligns perfectly with the purpose of our work.

\section{Ablation Study}
\subsection{Anomalous Images Generation}
\begin{figure*}[t]
    \begin{minipage}{.3\linewidth}
        \centering
        \includegraphics[width=1.0\textwidth]{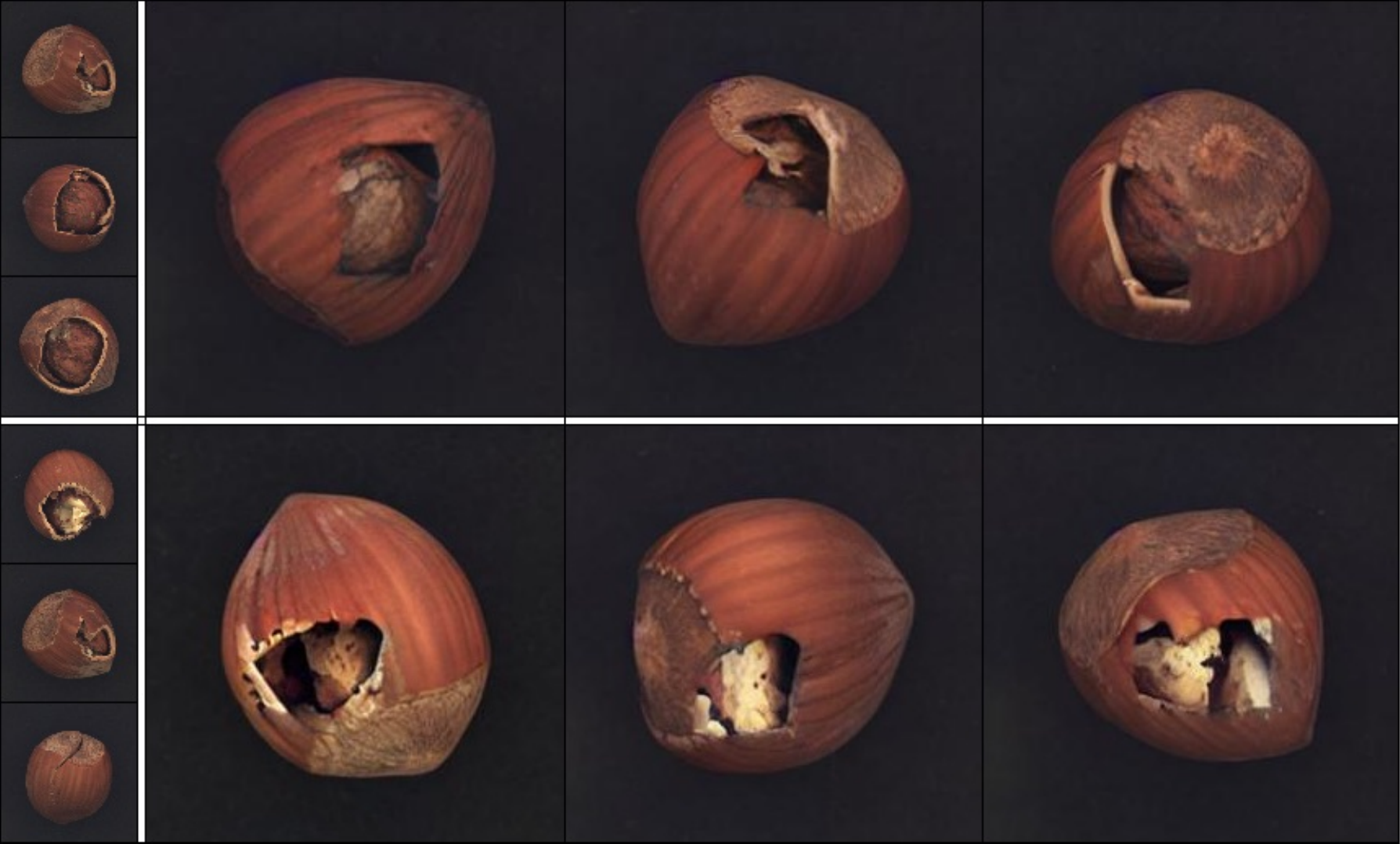}
        \subcaption*{(a)}
        \label{fig4a}
    \end{minipage}
    \hfill
        \begin{minipage}{.33\linewidth}
        \centering
        \includegraphics[width=1.0\textwidth]{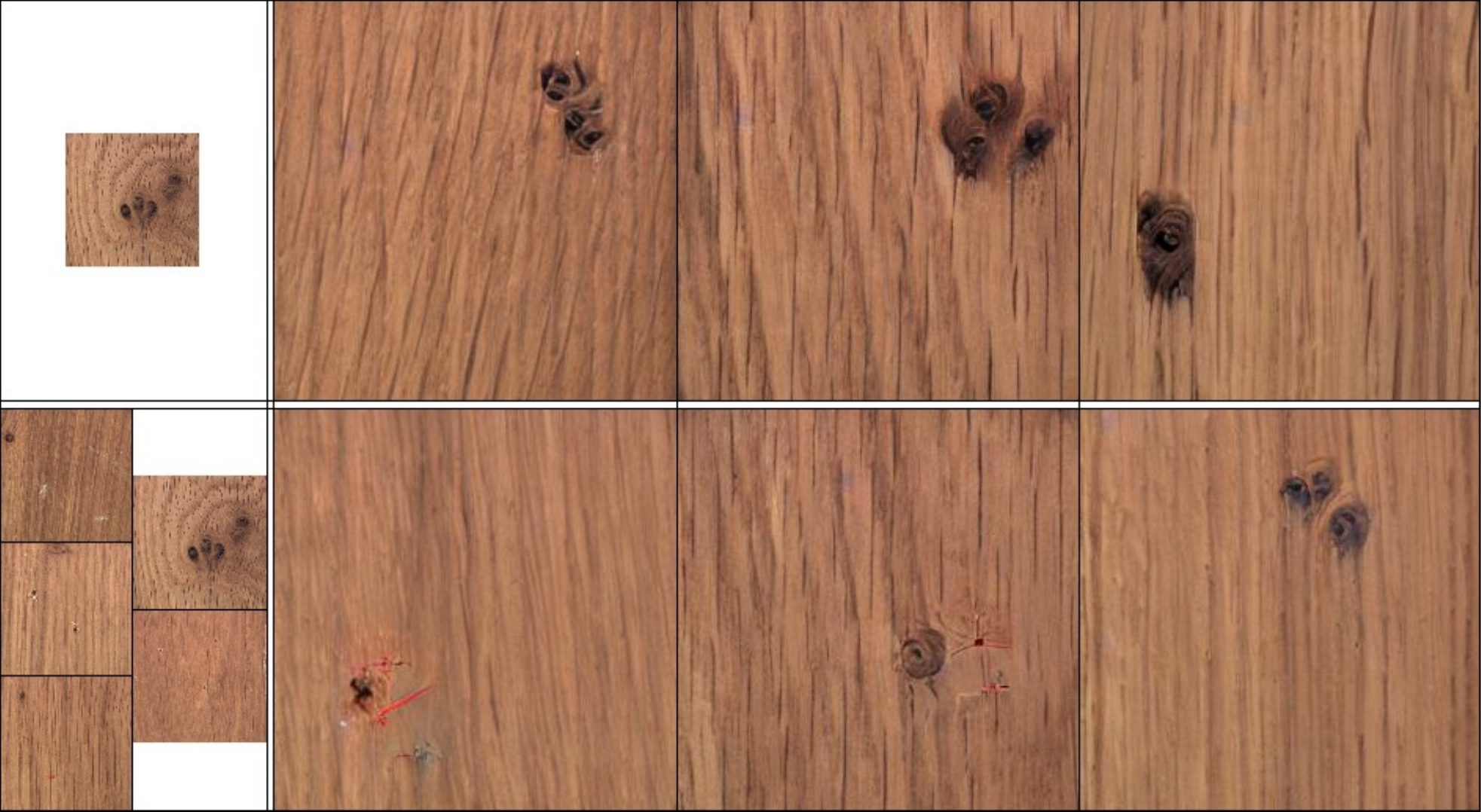}
        \subcaption*{(b)}
        \label{fig4b}
    \end{minipage}
    \hfill
        \begin{minipage}{.3\linewidth}
        \centering
        \includegraphics[width=1.0\textwidth]{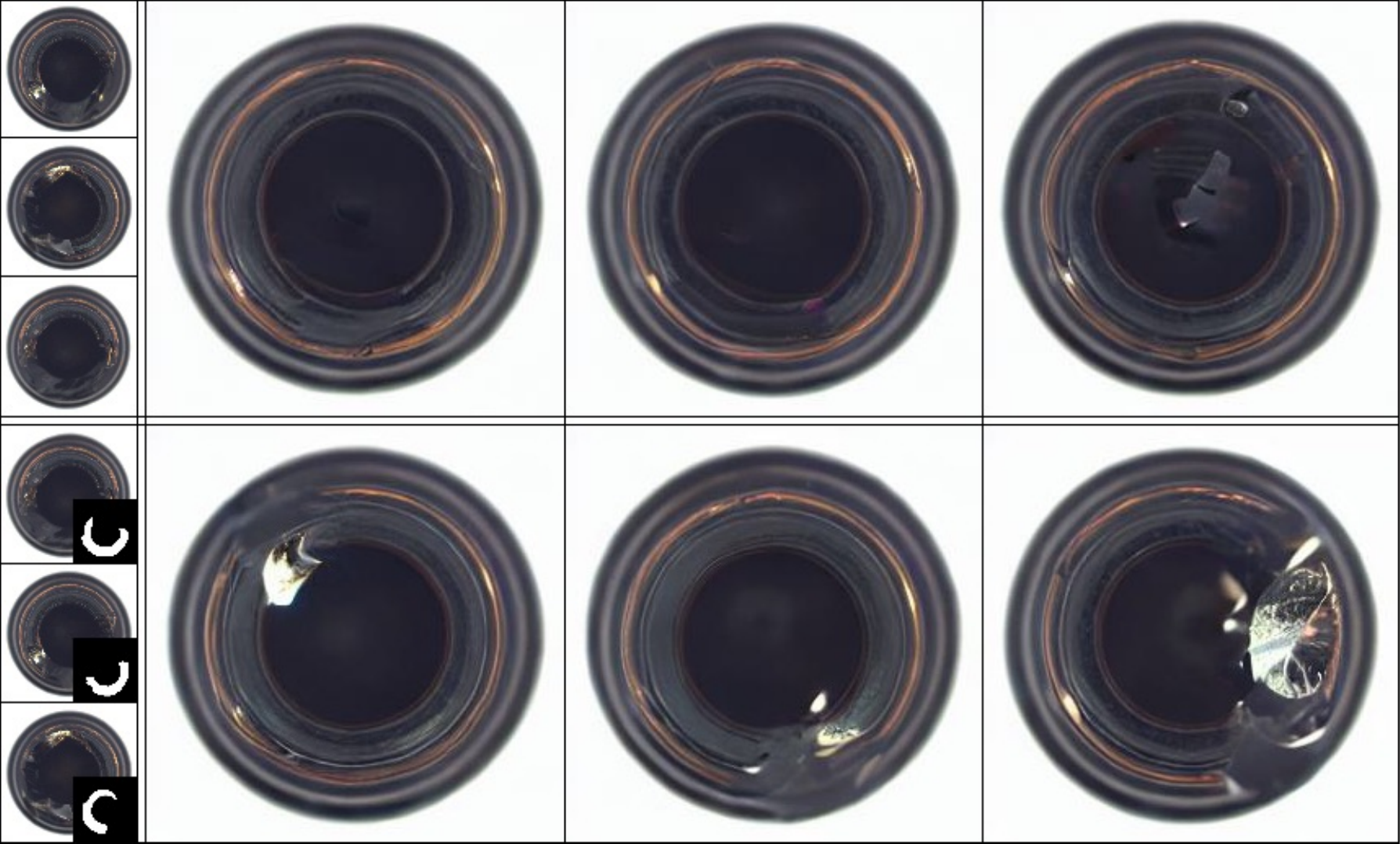}
        \subcaption*{(c)}
        \label{fig4c}
    \end{minipage}
    \caption{The ablation study (visualization of the generated images). 
    (a) Comparison of different support anomalies, the number of images is fixed to 3. (b) Comparison of different numbers of support anomalies. (c) Comparison between mask guide loss and non-mask guide loss during embedding learning.
    }
    \label{fig4}
\end{figure*}

\noindent\textbf{Effects of different support anomalies}. We analyze the impact of different support anomalies, and the results are presented in Figure~\ref{fig4}\textcolor{red}{a} and Table~\ref{tab2a}.
In Figure 4 (a), both sets of abnormal images are associated with the object of ``\textit{crack hazelnut}''. However, the instances in the support set are inconsistent, with one set (the second row) revealing the \textit{white kernel}, causing the generated images to also tend to reveal the \textit{white kernel}. Despite this inconsistency in instances, the semantics (``\textit{crack hazelnut}'') remain consistent and correct. As a result, the impact on model performance is minimal, as shown in Table~\ref{tab2a}.
This indicates that different support sets can affect the generated instances, but the semantics remain aligned with the intended expectations. 

\begin{table*}[t]
\renewcommand\arraystretch{0.9}
    \centering
    \tiny
        \caption{The ablation study (model performance on anomaly detection task). 
        }
    \vspace{-5pt}
    \begin{subtable}[t]{0.32\textwidth}
    \caption{Effects of different support anomalies.}
    \vspace{-2pt}
        \begin{tabular}{@{}cc|cc@{}}
\toprule
\multicolumn{2}{c|}{support set}                         & 1-st set     & 2-nd set     \\ \midrule
\multicolumn{1}{c|}{\multirow{2}{*}{Image}} & AU-ROC & 98.7 & 98.5\\
\multicolumn{1}{c|}{}                       & AU-PR  & 99.5 & 99.1 \\ \midrule
\multicolumn{1}{c|}{\multirow{2}{*}{Pixel}} & AU-ROC & 98.1 & 97.7 \\
\multicolumn{1}{c|}{}                       & AU-PR  & 73.2 & 71.4 \\
\bottomrule
\end{tabular}
        %\caption{}
        \label{tab2a}
    \end{subtable}
    \hfill
    \begin{subtable}[t]{0.32\textwidth}
    \caption{Effects of the number of support anomalies.}
    \vspace{-2pt}
        \begin{tabular}{@{}cc|ccc@{}}
\toprule
\multicolumn{2}{c|}{k-shot}                         & 1     & 3     &5\\ \midrule
\multicolumn{1}{c|}{\multirow{2}{*}{Image}} & AU-ROC & 97.7 & 98.7 & 98.6\\
\multicolumn{1}{c|}{}                       & AU-PR  & 98.9 & 99.5 & 99.5\\ \midrule
\multicolumn{1}{c|}{\multirow{2}{*}{Pixel}} & AU-ROC & 97.6 & 98.1 & 98.2\\
\multicolumn{1}{c|}{}                       & AU-PR  & 70.5. & 73.2 & 73.0\\
\bottomrule
\end{tabular}
        %\caption{}
        \label{tab2b}
    \end{subtable}
    \hfill
    \begin{subtable}[t]{0.32\textwidth}
    \caption{Effects of mask-guided learning loss.}
    \vspace{-2pt}
        \begin{tabular}{@{}cc|cc@{}}
\toprule
\multicolumn{2}{c|}{leaning embedding}                         & mask       & non-mask      \\ \midrule
\multicolumn{1}{c|}{\multirow{2}{*}{Image}} & AU-ROC & 98.7 & 97.8 \\
\multicolumn{1}{c|}{}                       & AU-PR  & 99.5 & 98.7 \\ \midrule
\multicolumn{1}{c|}{\multirow{2}{*}{Pixel}} & AU-ROC & 98.1 & 97.6 \\
\multicolumn{1}{c|}{}                       & AU-PR  & 73.2 & 70.4 \\
\bottomrule
\end{tabular}
        %\caption{}
        \label{tab2c}
    \end{subtable}
    \label{tab:array}
\end{table*}

\noindent\textbf{Effects of the number of support anomalies.} We also analyzed to assess the impact of different numbers of support anomalies. As depicted in Figure~\ref{fig4}\textcolor{red}{b}, when only one image is available, the learned distribution tends to be biased towards the specific features of that image. Consequently, the generated images lack diversity and generalization. This is reflected in Table~\ref{tab2b}, where the performance on anomaly detection tasks slightly decreases.
When using more images (5 images) to learn the embeddings, the diversity and generalization of the generated images significantly improve, as confirmed by the specific metrics in Table~\ref{tab2b} for anomaly detection tasks.
Considering that the performance using 3 support anomalies is comparable to using 5 support anomalies (0.1\% difference on image AU-ROC and 0.2\% difference on pixel AU-PR), we have opted for 3 images to reduce the demand for anomaly data.

\begin{figure*}[t]
\begin{minipage}{.40\textwidth}
    \renewcommand\arraystretch{1.1}
        \centering
        \scriptsize
                    \captionof{table}{Ablation study of $\tau$.}
            \begin{tabular}{@{}cc|cccc@{}}
            \toprule
            \multicolumn{2}{c|}{$\tau$}                         & 1.0     & 0.95   &0.90 &0.80  \\ \midrule
            \multicolumn{1}{c|}{\multirow{2}{*}{Image}} & AU-ROC & 98.5 & 98.5 & 98.7 & 98.2\\ 
            \multicolumn{1}{c|}{}                       & AU-PR  & 99.5 & 99.4 & 99.5 & 99.2\\ \midrule
            \multicolumn{1}{c|}{\multirow{2}{*}{Pixel}} & AU-ROC & 98.0 & 98.2 & 98.1 & 97.1\\
            \multicolumn{1}{c|}{}                       & AU-PR  & 68.9 & 71.1 & 73.2 & 65.4\\
            \bottomrule
            \end{tabular}
            \vspace{-6pt}
        \label{tab3}
    \end{minipage}
    \hfill
    \begin{minipage}{0.58\textwidth}
        \centering
        \scriptsize
                    \captionof{table}{Ablation study of  anomalies.}
            \begin{tabular}{l|ccc|c}
            \toprule
            Anomalies & CutPaste & DRAEM & Ours & AU-PR\\ \midrule
            DRAEM\textsubscript{CutPaste} & \ding{51} & & & 59.8 \\
            DRAEM\textsubscript{Original} & & \ding{51} & & 67.4 \\
            DRAEM\textsubscript{Ours} &&& \ding{51}& 68.3\\
            DRAEM\textsubscript{CutPaste+Ours} &\ding{51}&&\ding{51}&69.0\\
            DRAEM\textsubscript{DRAEM+Ours} &&\ding{51}&\ding{51}&73.2\\ 
            \bottomrule
            \end{tabular}
            \vspace{-5pt}
        \label{tab4}
    \end{minipage}
\end{figure*}

\noindent\textbf{Mask-Guided embedding learning}. 
We emphasize the importance of incorporating the ground truth of support anomalies when learning embeddings.
This is to ensure that the learning target is biased towards the anomaly region rather than the object (or texture). 
As shown in Figure~\ref{fig4}\textcolor{red}{c}, without the guidance of the mask, the distribution represented by the embeddings is biased towards the entire object (``\textit{bottle}''), failing to generate anomaly (``\textit{broken}''). At the same time, the performance of anomaly detection tasks is also severely affected.

\subsection{Anomaly Detection with Generated Anomalies}
\textbf{Confidence threshold $\tau$}. 
In the weakly supervised anomaly detection model, we use a threshold $\tau$ to filter out interference from high-confidence normal pixels within the bounding box.
As shown in Table \ref{tab3},  $\tau$ has a significant impact on segmentation tasks. 
For example, when the $\tau=0.9$, the pixel-level AU-PR reaches 73.2\%, whereas when $\tau=1.0$ and $\tau=0.8$, the performance decreases to 68.9\% and 65.4\%, respectively.
This is because if the threshold is too low, the model will be forced to classify more normal pixels within the box region as anomaly pixels. Conversely, if the threshold is too high, the model will ignore learning anomalous pixels.  
Based on our experimentation, we tend to prefer setting the threshold to 0.9 or 0.95, as it strikes a balance between capturing anomalous pixels and avoiding misclassification of normal pixels within the bounding box.

\noindent\textbf{Training with different anomalies}. We also investigate the impact of the training model with different anomalies. 
As shown in Table~\ref{tab4}, both DRAEM-based synthetic anomalies and our generated anomalies can achieve good performance on pixel AU-PR.
The performance of the model trained solely on DRAEM-based anomalies or our anomalies is comparable (67.4\% and 68.3\%, respectively). Furthermore, when these anomalies are used together, a significant improvement can be observed, reaching 73.2\%
This can be attributed to the fact that support anomalies may not fully represent the distribution of the test set, resulting in weaker predictive capabilities for unseen images. On the other hand, DRAEM-based anomalies, which incorporate a large amount of out-of-distribution data, can enhance the model's ability to predict unseen images. Therefore, training the model jointly on both DRAEM-based anomalies and our anomalies proves to be a better choice.

\begin{wrapfigure}{r}{0.45\textwidth}%[htbp]
    \centering
    \includegraphics[width=0.45\textwidth]{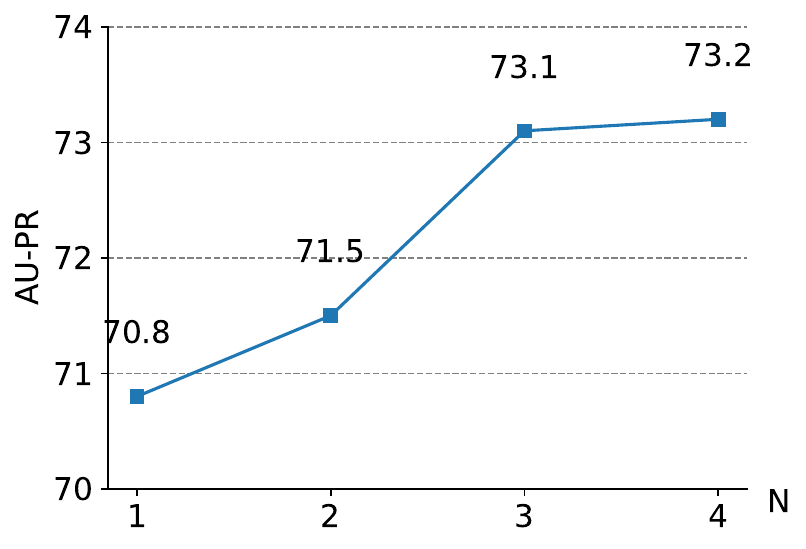}
    \caption{Ablation study of $N$.}
    \label{fig5}
    \vspace{-10pt}
\end{wrapfigure}

\noindent\textbf{The number of generated anomaly images}. In our study, we generate $N=4$ for each object (or texture), and Figure~\ref{fig5} illustrates the impact on the model with different values of $N$.
As $N$ increases, the model benefits from having more anomalies. When $N<3$, it is evident that increasing the number of anomalies is beneficial for the model. 
However, it is important to consider that generating more anomalies requires additional time and resources.
Taking into account the trade-off between model performance and the cost of image generation, we ultimately chose $N=4$ as the optimal value.

%% file: sections/6_conclusion.tex
\section{Conclusion}
In this work, we have identified and addressed two key challenges in anomaly detection. 
The first challenge is the scarcity of available real-world abnormal images, which makes it difficult to train anomaly detection models effectively. 
The second challenge is that synthetic anomalies used in previous methods are unrealistic and have a significant semantic gap compared to real-world anomalies. 
To address these challenges, we propose an anomaly-driven generation method (AnoGen) to generate a large number of real and diverse abnormal images. 
Since real abnormal images are scarce, our generation process is driven by few-shot learning, requiring only three real abnormal images. 
By applying these generated images to DRAEM and DeSTSeg, we achieved promising results on the widely used MVTec anomaly detection dataset. 
In particular, our generated images have significantly improved the performance on anomaly segmentation task, as evidenced by the AU-PR metrics. This demonstrates the effectiveness of our approach in addressing the above challenges in anomaly detection and highlights the potential of our method for enhancing anomaly detection in real-world scenarios.
Overall, our approach uses a small number of real anomaly images to significantly improve model performance, and since a small number of real anomaly images are present in many real-world scenarios, this is of great significance for many practical applications of industrial anomaly detection.

\noindent \textbf{Limitation.} 
The generated anomaly images enjoy free annotations but in bounding boxes form. Therefore, pixel-level model training needs the efforts of weakly supervised methods, which inevitably introduces additional hyperparameters. In our future work, we will explore how to obtain more precise annotations when generating anomaly images so that they can be more easily and widely used in anomaly detection.